\documentclass[review]{interact}
\usepackage[]{setspace}
\usepackage{epstopdf}
\usepackage{subfigure}
\usepackage[T1]{fontenc}
\usepackage{url}
\usepackage{hyperref}
\usepackage{soul}
\usepackage{natbib}
\bibpunct[, ]{(}{)}{;}{a}{}{,}
\renewcommand\bibfont{\fontsize{10}{12}\selectfont}

\theoremstyle{plain}

\theoremstyle{definition}

\theoremstyle{remark}

\begin{document}

\title{Analytics and Machine Learning in Vehicle Routing Research}
\author{
\name{Ruibin Bai\textsuperscript{a}\thanks{The authors are presented in alphabetic order. Contact R. Bai Email: ruibin.bai@nottingham.edu.cn for correspondence.} and Xinan Chen\textsuperscript{a} and Zhi-Long Chen\textsuperscript{b} and Tianxiang Cui\textsuperscript{a} and Shuhui Gong\textsuperscript{c} and Wentao He\textsuperscript{a} and Xiaoping Jiang\textsuperscript{d} and Huan Jin\textsuperscript{a} and Jiahuan Jin\textsuperscript{a} and Graham Kendall\textsuperscript{e,f} and Jiawei Li\textsuperscript{a} and Zheng Lu\textsuperscript{a} and Jianfeng Ren\textsuperscript{a} and Paul Weng\textsuperscript{g,h} and Ning Xue\textsuperscript{i} and Huayan Zhang\textsuperscript{a}}
\affil{\textsuperscript{a} School of Computer Science, University of Nottingham Ningbo China, Ningbo, China.  \\
\textsuperscript{b} Robert H. Smith School of Business, University of Maryland, MD 20742, USA. \\ 
\textsuperscript{c} China University of Geosciences, Beijing, China.\\
\textsuperscript{d} National University of Defence Technology, Hefei, China\\
\textsuperscript{e} School of Computer Science, University of Nottingham, UK.\\
\textsuperscript{f} School of Computer Science, University of Nottingham Malaysia, Malaysia\\
\textsuperscript{g} UM-SJTU Joint Institute, Shanghai Jiao Tong University, Shanghai, China.\\
\textsuperscript{h} Department of Automation, Shanghai Jiao Tong University, Shanghai, China.\\
\textsuperscript{i} Faculty of Medicine and Health Sciences, University of Nottingham, UK.}
}


\date{May 2020}

\maketitle
\begin{abstract}
The Vehicle Routing Problem (VRP) is one of the most intensively studied combinatorial optimisation problems for which numerous models and algorithms have been proposed. To tackle the complexities,  uncertainties and dynamics involved in real-world VRP applications, Machine Learning (ML) methods have been used in combination with analytical approaches to enhance problem formulations and algorithmic performance across different problem solving scenarios. However, the relevant papers are scattered in several traditional research fields with very different, sometimes confusing, terminologies. This paper presents a first, comprehensive review of hybrid methods that combine analytical techniques  with  ML tools in addressing VRP problems.  Specifically, we review the emerging research streams on ML-assisted VRP modelling and ML-assisted VRP optimisation. We conclude that ML can be beneficial in enhancing VRP modelling, and improving the performance of algorithms for both online and offline VRP optimisations.   Finally,  challenges and future opportunities of VRP research are discussed. 
\end{abstract}


\section{Background and Motivation} 
The Vehicle Routing Problem (VRP) is one of the most studied problems in the field of operations research.  A search using keyword ``vehicle routing'' on Clarivate's Web of Science returns more than 8,000 papers, including 131 review papers. One reason for this significant research attention is due to the booming e-commerce industry that leads to exponential growth in transportation and logistics. With the advances in computing power and progresses in modelling and solution methodologies, it is now possible to solve VRPs of much larger sizes in less time than we could in the past. 
There have been a number of survey papers related to VRP. \cite{vidal_concise_2020} provided a good overview of different VRP variants, including the emerging variants characterised by different objectives and performance metrics.  \cite{braysy_vehicle_2005,braysy_vehicle_2005-1} conducted comprehensive reviews on the heuristic methods for different VRPs. Most of the papers they reviewed focus on deterministic VRPs, in which the problem parameters are assumed to be deterministic and known prior to the problem solving. \cite{gendreau_stochastic_1996} provided a review on stochastic vehicle routing in which some of the problem parameters are assumed stochastic, while \cite{pillac_review_2013} surveyed all the dynamic vehicle routing problems in which the problem parameters are revealed dynamically over time. Given their close relevancy between stochastic VRP and dynamic VRP, \cite{ritzinger_survey_2016} provided a combined review for both the dynamic and stochastic vehicle routing problems. 

Although a tremendous amount of research has been devoted to VRP problems, it is still very difficult to tackle some practical VRP applications for the following reasons.  
Firstly, the majority of existing VRP research focuses on the analytical properties of different VRP variants and the corresponding solution methods. This type of research is often dominated by the use of mathematical models to define key objectives and constraints \citep{vidal_concise_2020}. However, for the convenience of theoretical analyses and problem solving, almost all mathematical models are associated with a number of assumptions, some of which may not be practical for real-life applications. It can also be challenging to estimate relevant problem parameters. For practitioners, it becomes a hurdle in translating the existing models and algorithms into successful real-life applications. 

Secondly, numerous VRP models have been developed to mathematically formulate uncertainties pertaining to the VRPs. However, most of them are limited to theoretical or small-scale empirical studies. Implementation of these models and the proposed solution methods in real-world applications is rare and still faces considerable challenges. There is a growing demand for making these models more practically applicable. 

To tackle some of these issues, such as unrealistic model assumptions, difficulties in parameter estimation and the practicality of solution algorithms, there is an emerging VRP research direction of using hybrid methods that combine data analytics and machine learning tools with conventional optimisation based techniques.  With the assistance of analytics and ML, conventional VRP modelling and solution techniques can be significantly strengthened. This paper aims to provide a comprehensive review of such hybrid methods for VRP applications. 

VRP research is traditionally limited to the Operations Research (OR) community. However, with the advances in machine learning methodologies, researchers in this community have recently made attempts to tackle combinatorial optimisation problems (including VRPs) solely using machine learning methods (i.e. without explicitly exploiting the structures of the mathematical models). These methods often, despite some progress, suffer from issues such as the lack of generalisation across different scenarios, inefficiency in data use, and the inability to discover insights and interpret solution structures. There seems to be very little interaction between the two communities. Related papers are scattered in journals of both research communities and cross-community paper citations are fewer than you would expect. This leads to the lagged acknowledgement of progress made across research communities. Furthermore, each community uses its own set of terminologies such that similar ideas are defined with different terms. This often causes considerable confusion for researchers and practitioners. We believe that a thorough review of the VRP research that uses tools from these two research communities would be useful to both communities. 

Industrial partners also require a holistic review of  existing VRP modelling techniques that explain the practicability of these formulations in terms of quality, availability of data and how parameters can be estimated with required precision in order to satisfy the assumptions and other engineering requirements. In this review, we include a section to existing research studies on how machine learning has been used in achieving more practical VRP modelling and parameter estimation.  We will also review existing studies that use ML to help build more efficient algorithms for VRP problems.  

This review paper is not intended to give another comprehensive review of all VRP-related papers. Instead, the focus is on a new, fast-growing VRP research direction that investigates novel ways to integrate existing analytical methods based on mathematical models with advanced ML methodologies.

We recognise the long history of such research efforts in addressing VRP problems, but it has drawn particular research attention in the past three years, partly due to the growing popularity of analytics and ML research in the OR community, and partly due to a shift of focus in the ML community from single node intelligence to complex system intelligence. We believe that, to achieve major breakthroughs to realise full system intelligence, researchers from both the OR community and the ML community need to collaborate at a much deeper level to tackle the significant challenges that VRP presents. This review paper aims to bridge the gap and promote more interactions and collaborations between the two communities. We expect that our review will inspire more researchers to tackle complex VRP problems and make a positive impact on our daily life. 

\begin{figure}
    \centering
    \includegraphics[scale=0.8]{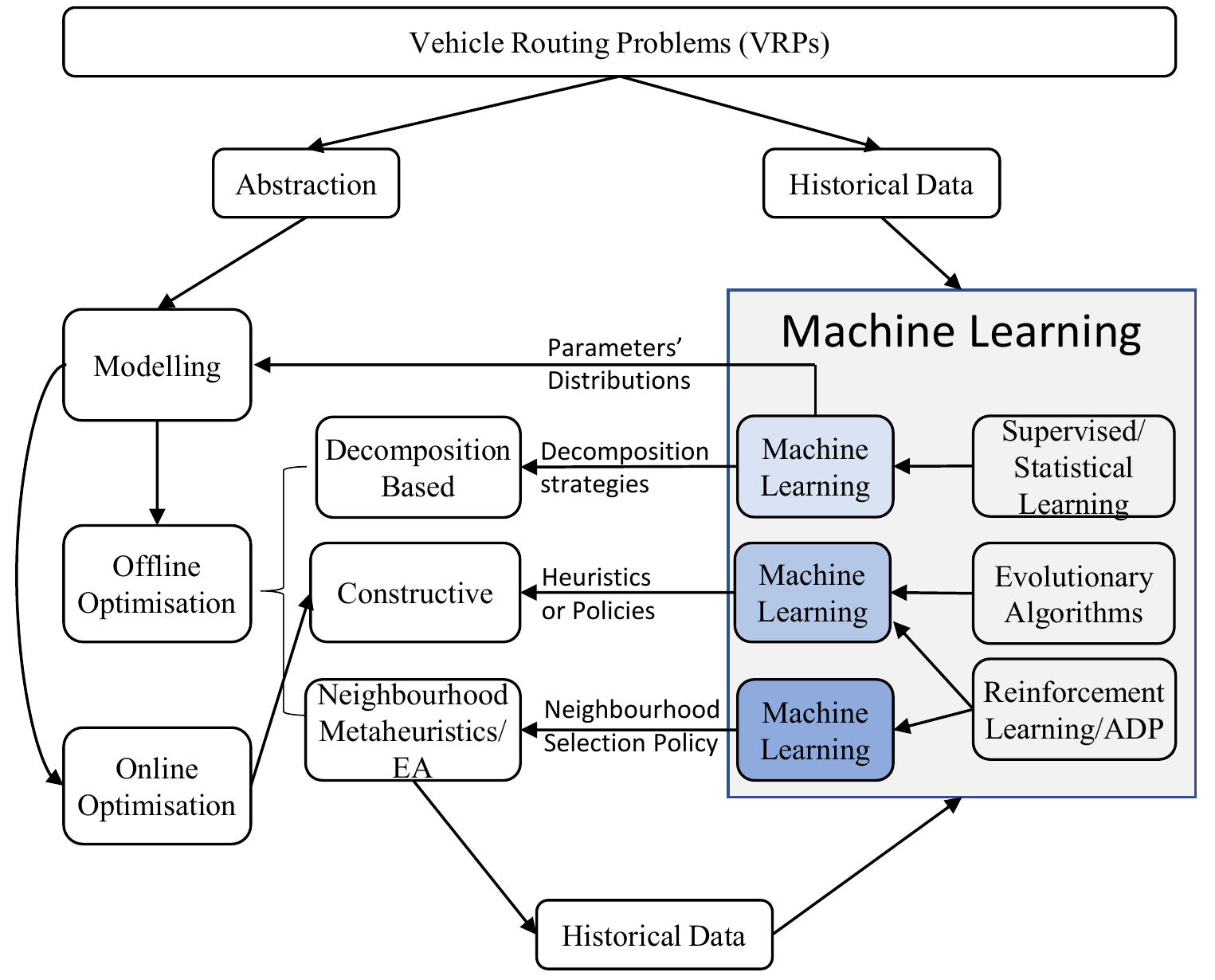}
    \caption{The proposed classifications of machine learning assisted VRP research.}
    \label{fig:classify}
\end{figure}

Figure \ref{fig:classify} illustrates the overall classification of related research papers that hybridise ML with analytical approaches in the VRP. We broadly classify three major types of integration efforts. They are ML-assisted VRP modelling, ML-assisted offline and ML-assisted online optimisations, respectively. Most machine learning methodologies require historical data of some problem parameters. Some of the methods will also require meta-data generated by the optimisation methodologies.

The remainder of this review is organised as follows: 
In Section~\ref{sec2}, we provide an introduction to VRP problems, its main variants and the main algorithms that have been utilized. Section~\ref{sec3} reviews modelling methodologies for VRPs with uncertainties, including stochastic programming, robust optimisation, chance constrained programming, and data analytics and forecast. In Section~\ref{sec4}, ML-assisted VRP algorithms are reviewed according to how the machine learning is utilised in the solution methods, including decomposition based methods, adaptive neighbourhood search and machine learning trainable constructive methodologiess. Section~\ref{sec5} provides some concluding remarks and discusses the challenges, prospects and opportunities for the future research.

\section{Introduction} \label{sec2} 
    For the benefit of those new to VRP, in this section, we give a general introduction of VRP problem, its main variants, and commonly used algorithms.  
    \subsection{Basic Vehicle Routing Problem}
    
    
    
    
    The basic VRP, proposed by~\cite{dantzig1959truck}, can be defined as an optimisation problem comprising of a set of distributed customers, each with a freight demand, and a fleet of vehicles starting from the central depot. The objective is to find minimal travel cost (e.g. distance or time), such that each customer is visited and served by a vehicle exactly once. The VRP is related to one of the most extensively studied combinatorial optimisation problems, the Travelling Salesman Problem (TSP), which was first considered by~\cite{menger1932botenproblem}. While TSP aims to find a circular shortest path to traverse all customers without consideration of several practical constraints related to capacity and time, the basic VRP problem also takes into account the capacity constraints related to vehicles and hence requires to find multiple vehicle routes with the minimum total cost. The basic VRP problem is also called the Capacitated VRP (CVRP). For a given set of customers $V$ and vehicle depot node 0, a commonly used mathematical model for CVRP is the following vehicle flow formulation:
    
    \begin{equation}\label{cvrp}
        \min \sum_{i\in V}{\sum_{j\in V}c_{ij}x_{ij}}
    \end{equation}
    subject to
    \begin{subequations} 
    \begin{align}
        \sum_{i\in V}{x_{ij}} &= 1 \quad \forall j\in V \setminus\{0\} \\
        \sum_{j\in V}{x_{ij}}&=1 \quad \forall i\in V \setminus\{0\} \\
        \sum_{i\in V}{x_{i0}}&=K  \\
        \sum_{j\in V}{x_{0j}}&=K  \\
        \sum_{i\notin S}{\sum_{j\in S}x_{ij}} &\geq r(S) \quad \forall S\subseteq{V}\setminus \{0\}, S\neq \emptyset
  \end{align}
\end{subequations}
    where $x_{ij}$ is a binary decision variable that indicates whether the arc $(i,j)$ is part of the solution and $c_{ij}$ is the cost of using arc $(i,j)$. $K$ is the number of vehicles being used and $r(S)$ is the minimum number of vehicles required to serve customer set $S$. Constraints (2a,2b) make sure that each customer is visited exactly once and constraints (2c,2d) ensure the satisfaction of the number of vehicle routes. Finally constraint (2e) makes sure that the demands from all customers are fully satisfied. VRP is a classical NP-hard problem, thus most solution algorithms are heuristic-based.

    \subsection{Main VRP Variants}

    Due to the complexities in real-world VRP problems, the basic VRP problem has been extended into a number of variants. The review work by \cite{braekers2016vehicle} refers to a taxonomy from \cite{eksioglu2009vehicle}, which provides a detailed categorisation of various VRP problems and their models. They classified VRPs with different constraints and objectives as `scenario characteristics', and VRP for different real-world applications as `problem physical characteristics'. They also defined an `information characteristics' category and a `data characteristics' dimension, based on the nature of the problem data. 
    
    The focus of this paper is mainly on the potential benefits of using data analytics and machine learning for better VRP formulations and problem solving. Specifically, for some conventional VRP variants, machine learning models can potentially be applied as probabilistic modelling technique for customer arrivals, demand quantities, waiting times, etc. For example, when vehicles are constrained to deliver services to customers within certain time intervals, it is known as the Vehicle Routing Problem with Time Windows (VRPTW). If a number of goods are transported from certain pickup locations to other delivery locations, it is known as the Pickup and Delivery Problem (PDP). If part of the problem components remains uncertain and follows a probability distribution, it is defined as Stochastic Vehicle Routing Problem (SVRP). For different VRP modelling, machine learning methodologies can be implemented in  historical data analysis to get the prior knowledge. We adopt the taxonomy proposed in \cite{eksioglu2009vehicle} to highlight the VRP variants for which machine learning techniques could be used to enhance the practicality and quality of the problem modelling.  
    
    \begin{enumerate}
        \item {\bf Scenario Characteristics}
        \begin{itemize}
            \item {\bf Number of stops/customers} on a route is partially known or partially probabilistic \citep{albareda2014dynamic,snoeck_route_2020}.
            \item {\bf Customer service demand quantity} is stochastic or unknown. \citep{zhang2013single,dinh2018exact,ghosal2020distributionally,markovic2005using}. 
            \item {\bf Request times} of new customers are stochastic or unknown. 
            \item \textbf{Onsite service/waiting times} are stochastic or unknown. \citep{li2010vehicle,zhang2013stochastic}.
        \end{itemize}
        \item {\bf Problem Physical Characteristics}
        \begin{itemize}
            \item {\bf Time window restrictions} (on customers, roads, and at facilities) are stochastic or unknown \citep{chuah_routing_2005}. 
            \item \textbf{Travel time} is stochastic or unknown.  \citep{musolino2013travel,han2014robust,li2010vehicle,tacs2013vehicle,tacs2014vehicle}.
        \end{itemize}
        \item Information Characteristics
        \begin{itemize}
            \item \textbf{Evolution of information} is partially dynamic.  \citep{sabar2014dynamic}.
            \item \textbf{Quality of information} is stochastic or unknown. \citep{balaji_orl_2019}.
            \item \textbf{Availability of information} is local to only a subset of participants concerned in the problem. 
            
        \end{itemize}
        \item Data Characteristics
        \begin{itemize}

            \item \textbf{Data used} is from the real world which can be noisy and messy.  \citep{markovic2005using,bent2005online,li2018data,vzunic2020adaptive}
        \end{itemize}
    \end{enumerate}
    
Another VRP variant is the automated guided vehicle (AGV) routing, which is essentially a driver-less transport system used in logistics warehouses and marine container terminals \citep{vis2006survey}. Because of the availability of advanced communication and control mechanisms in real-time, AGV routing problems are often modelled as a dynamic problem \citep{qiu2002scheduling}. As a result,  data analytical methods and machine learning play a major role in tackling these type of problems as an improved version over simple heuristic based methods (e.g. \citep{grunow2006strategies,wang2015path,zhang2019development}).  Readers can refer to detailed surveys \citep{qiu2002scheduling,vis2006survey,hasan2019automated} for better understanding of AGV routing.


    
    \subsection{Types of VRP algorithms}
    Considering that the VRP and its variants belong to the NP-hard class of problems, exact algorithms are only applicable under certain circumstances and for small-scale problems. Exact methods treat VRP as integer or mixed-integer programs, and try to find a (near-)optimal solution. Since real-life VRPs are usually of large sizes, heuristic-based methods are considered more suitable. \cite{elshaer2020taxonomic} states that more than 70\% of solution methods in the literature are based on metaheuristics, which have various ``meta'' strategies capable of escaping from poor local optima but cannot guarantee optimality \citep{boussaid2013survey}. Representative exact algorithms for VRP include branch-and-price algorithm \citep{christiansen2007branch} and branch-and-cut algorithms \citep{augerat1995computational, ralphs2003capacitated, baldacci2004exact}. Metaheuristics can be divided into two categories: single-point based heuristics, and population-based heuristics. The former includes classical heuristic methods, such as simulated annealing (SA) \citep{kirkpatrick1983optimization}, tabu search (TS) \citep{glover_tabu_1997}, GRASP (for \textit{greedy randomized adaptive search procedure}) \citep{feo1995greedy}, variable neighborhood search (VNS) \citep{mladenovic1997variable}, guided local search (GLS) \citep{voudouris2003guided}, iterated Local Search (ILS) \citep{stutzle1999local}, large neighborhood search (LNS) and adaptive large neighborhood search (ALNS)~\cite{pisinger2010large}. The latter includes two  main types, evolutionary algorithms and swarm intelligence, which are mainly inspired by some natural phenomena. The readers may refer to \cite{elshaer2020taxonomic} for a comprehensive review on metaheuristic for VRP. This review focuses on the machine learning assisted VRP algorithms and Section \ref{sec4} discusses all the relevant papers.

\section{ML-Assisted VRP Modelling} \label{sec3} 
This section provides an overview of various VRP modelling methodologies that are supported by data analytics and machine learning. In particular, we focus on the modelling techniques for handling uncertain, incomplete, imprecise or ambiguous data in VRPs, including stochastic programming, robust optimisation, chance constrained programming and data forecast. 

\subsection{Stochastic Programming for VRP Modelling}

Machine learning has long been used to assist the modelling of stochastic VRPs, where the uncertain data is represented by random variables with known probability distributions. A number of studies have modelled stochastic VRPs as a Markov Decision Process (MDP), which can be tackled by Neuro-Dynamic Programming (NDP). NDP is also referred to as reinforcement learning in the artificial intelligence literature \citep{bertsekas_neuro-dynamic_1996}. The value-function approximation method and/or the policy-function approximation method of NDP is tailored in each of those studies. \cite{secomandi_comparing_2000} compared two different NDP algorithms on solving VRPs with stochastic demands and showed that the NDP with rollout policy performs better than an approximate policy iteration.
\cite{godfrey_adaptive_2002} formulated the dynamic fleet management 
problem as a multistage dynamic program and use gradient-based information to obtain nonlinear approximations of value functions.
\cite{zhang2013single} used tabular value approximation 
and policy approximation to solve VRPs with stochastic demands.
\cite{ulmer2020modeling} presented a route-based Markov decision process modelling framework for dynamic VRP that extends the conventional MDP model for dynamic and stochastic optimisation problems by redefining the conventional action space to operate on route plans. The proposed modelling framework makes it easier to integrate machine learning approaches with the existing route-based analytical methods. 

Machine learning has also been used to estimate the probability distributions of uncertain data in stochastic VRPs. 
\cite{talbi_hybrid_2013} state that hybrid metaheuristics are often used to solve complex and large real-world optimisation problems, combining advantages from machine learning techniques and mathematical optimisation. The authors examined the hybrid metaheuristics for dynamic stochastic VRP, where not all information is available in advance.
\cite{bent2005online} proposed a machine learning approach by sampling 
from historical data to get partial knowledge of stochastic distribution in online scheduling problems. They claim that their approach could help widen the application area of stochastic algorithms. Experiments are conducted on several problems including dynamic VRPTW from previous work \cite{bent_scenario-based_2004}.
\cite{defourny_machine_2010}  generated and selected scenario trees with random branching structures for multi-stage stochastic programming over large planning horizons. This work combines the perturb-and-combine estimation methods from machine learning with stochastic programming techniques for sequential decision making under uncertainty. The performance has been shown to be competitive. \cite{bai_stochastic_2014} studied a two-stage model for stochastic transportation network that supports vehicle rerouting in the second stage. 
    
\subsection{Robust Optimisation for VRP Modelling}
 Stochastic programming assumes that the probability distributions of uncertain parameters are known, and aims for the best average performance over all possible future scenarios. By contrast, robust optimisation assumes that the probability distributions of uncertain parameters can be unknown, but the range of the uncertain data is known, and aims for a solution that could work even in the worst-case scenario. In general, robust optimisation is considered as a risk-averse method. 
 
Robust CVRPs require the vehicle routes to be feasible for all customer demands within a pre-specified uncertainty set (e.g., a box, polyhedron, or ellipsoid). Branch-and-cut schemes for the exact solution of the robust CVRPs have been proposed by \citep{Ilgaz2008} and \citep{Gounaris2013}. Robust CVRPs are amenable to solution schemes that appear to scale better than those for the chance constrained CVRPs (see Section 3.3). However, the solutions obtained from the robust CVRPs can be overly conservative since all demand scenarios within the uncertainty set are treated as equally likely, and the routes are selected solely in view of the worst demand scenario from the uncertainty set(\citet{ghosal2020distributionally}). 

For a comprehensive review of robust optimisation, we refer interested readers to \citep{Bertsimas2010}.
More recent contributions are some simulation based approaches, which transform the complex stochastic VRPs into a set of deterministic CVRPs, for which fast and extensively tested meta-heuristics exist \citep{Marcella2020}. To our knowledge, very little, if any, research has been conducted so far to apply machine learning techniques to robust VRPs, which offers many research opportunities.

\subsection{Chance-constrained Programming for VRP Modelling}

Chance-constrained programs are different from the stochastic programming and robust optimisation.  It assumes that constraints are satisfied to some degree, measured by probabilities, rather than fully met. In VRPs, the chance-constrained CVRPs consider vehicle routes that satisfy the customer demands along each route with a pre-specified probability. 

\cite{dinh2018exact} modelled the stochastic demand of customers as a joint normal distribution or a given discrete distribution. 
In particular, such modelling allows correlation between customer demands rather than independent demands. A branch-and-cut-and-price (BCP) algorithm is used to tackle the problem. Route relaxation is proposed in order to price through dynamic programming. The experiments demonstrated this approach's ability to solve the distributionally robust chance constrained vehicle routing problems (CCVRPs). 

\cite{ghosal2020distributionally} focused on the distributionally robust chance-constrained CVRP, which assumes that the probability distribution of the customer demand is partially known. The customer demands are modelled as the distribution based on ambiguity sets. Several first-order and second-order moment ambiguity sets are investigated and branch-and-cut schemes are used to solve it. The experiments demonstrated the good scalability of distributionally robust CVRP. 

\cite{li2010vehicle} studied a version of VRP where the service and travel times are stochastic and time window constraints are involved. Such problem is then formulated as a chance constrained programming model and a stochastic programming model with recourse. The author assigned a confidence level for both time window and driving duration of each route. Tabu search is utilised as the algorithmic solver. Similarly, \cite{gutierrez2018multi} also focused on VRP with stochastic service and travel times and time windows. The arrival times are estimated by a log-normal approximation. A multi-population memetic algorithm (MPMA) is then proposed as the algorithm to address the problem. \cite{wu2020chance} studied VRP for application of wet waste collections. In this work, chance constrained programming is used to model the uncertainty of waste generation rate and make it deterministic. A hybrid algorithm consisting of Particle Swarm Optimisation (PSO) and Simulated Annealing (SA) are used for the optimisation.

Similar to the robust optimisation methods, very little research has been conducted to explicitly combine machine learning with chance-constrained programming in solving VRP problems. However, one obvious way to hybridise them is to use ML to estimate probabilities or the parameters of the distributions. Another hybridisation opportunity can be using ML to speed up the various branch and pricing methods developed for chance-constrained programming. 

\subsection{Data Analytics and Forecast in VRP Modelling} 

In this subsection, we discuss a relatively loose connection between machine learning and the VRP. Research works in this area are not likely to use machine learning to address the related problem, but use it as a tool to reflect and analyse data that is related to problem solutions. A rough classification by \cite{talbi2016combining} divides such combination into the low-level integration, which is highly related to hyperparameter tuning; and the high-level integration, where machine learning is applied to help the modelling and provide certain contexts for metaheuristics. In this section, we mainly focus on machine learning assisted predictive models for handling uncertainties in VRP, and machine learning based model fine-tuning methods.

\subsubsection{Predictive Models for Uncertainties in VRP}

So far, a great number of methods have been proposed to solve classic VRP and many of them can generate high-quality solutions. However, most of these methods fail to work in realistic scenarios because of the over-simplification of the real-world constraints and uncertainties. For example, the elements such as customer occurrences and demands, and travel time may not be deterministic. Therefore, more sophisticated approaches are required to model and solve these problems. 

\cite{ce2010solving} made a basic classification by analysing customer's demands and introduced corresponding data mining techniques for each case. 
\cite{markovic2005using} used a neural network to predict customers' stochastic demand based on historical data, while \cite{snoeck_route_2020} applied probabilistic graphical model to predict constrained customers for routing problems. \cite{musolino2013travel} developed a simulation-based method to forecast the travel time for an emergency vehicles routing problem. 
\cite{li_travel_2019} investigated various ensemble based machine learning methods to predict the travel time in vehicle routing applications. 

\cite{li2018data} applied a data-driven search algorithm to improve traditional metaheuristics methods for large-scale manufacturing logistic problems in real-world industry practice. In the VRP part, 
they learned a multinomial transition matrix by updating from historical expert data and generated valid solutions by sampling from the matrix. 

 \cite{calvet2016combining} considered a  realistic VRP setting where customers have different demands to different assigned depots. The matching between heterogeneous depots and customers with stochastic demand could significantly affect the profitability of the company. Thus, several regression models which estimate the customer demands based on their assigned depot are used to support the market segmentation strategy. 
 
 In dynamic VRP, a customer's availability indicates the available service time window of each customer, which is one of the important aspects that cause the dynamism in Dynamic VRP (DVRP). \cite{kucharska2019dynamic} focused on this dynamism and used algebraic-logical meta-model (ALMM) to solve dynamic VRP with predicted customer availability. 

Another commonly used modelling approach addressing uncertainties in VRP is  rolling horizon planning. Problems are solved iteratively over a planning horizon that shifts forward a constant or variable time span at each step. At each iteration, some of the uncertainties of the problem are estimated with historical data. \cite{cordeau2015rolling} used this mechanism for a dynamic multi-period auto-carrier transportation problem which considers balancing vehicle usage and demand in dynamic settings. \cite{wen2010dynamic} studied dynamic multi-period vehicle routing problem with multi-objective of minimising customer waiting time and total travel costs as well as balancing the daily workload. Historical data are used to estimate the average daily workload over the rolling horizon. 

Machine learning has also been applied to a wide variety of problems in transportation and logistics. \cite{zhang_data-driven_2011} reviewed various data-driven traffic management systems, in which computer vision and other learning methods for environment sense and prediction are adopted.  
\cite{zantalis_review_2019} presented a review on machine learning and IoT in smart transportation, while \cite{anuar_vehicle_2019} gave a review  on machine learning and VRP methods in humanitarian logistics.
 
\subsubsection{Learning to Configure Algorithms}
There are a great number of hyper-parameters in VRP algorithms such as acceptance criteria of simulated annealing algorithm or population size of family of nature-inspired algorithms. Usually, these hyper-parameters need to be carefully adjusted manually. Good default parameters could significantly affect the performance of algorithms \citep{bengio_machine_2020}. Finding such suitable parameters is generally called the parameter setting problem (PSP). Machine learning methodologies could also be used to tackle the PSP problem as data-driven approaches. The general idea is to train regression models that map the specific problem instance features to the algorithm parameters where the training data are sets of pre-adjusted parameters and the corresponding performance.

\cite{calvet2016statistical} proposed a novel statistical learning-based methodology for solving the PSP and used it to fine-tune the existing algorithm for multi-depot vehicle routing problem. \cite{kadaba_integration_1991} proposed a hybrid framework including machine learning and knowledge-based systems to solve routing and scheduling 
applications. Neural networks are used for model selection and GA hyper-parameters tuning. \cite{liu_machine_2017}
used simple clustering techniques to improve the parameter selection for GA.  \cite{vzunic2020adaptive} built a framework where the predictive models, such as support vector machine, were used to adaptively control the hyper-parameters for each instance based on the historical data. The performance of the framework was demonstrated by testing on heterogeneous fleet vehicle routing problem with time windows (HVRPTW) along with a set of realistic logistics constrains.  

Clustering methodologies are also used in VRP modelling. \cite{hu2007intelligent} proposed a novel framework for solving CVRPTW. In the first stage of this framework, customers are clustered into different groups based on their features. The experiments demonstrated that such clustering could efficiently support the subsequent stages to generate feasible solutions. \cite{costa2020cluster} used a Cluster-based Hyper-Heuristic (CbHH) that adaptively cluster the customers to determine the neighborhood searching space at each step. The genetic algorithm is used to evolve the sequence of low-level perturbative operator to improve the initial solution. The experiment showed the effectiveness of the clustering technique and GA-based heuristic search. Similarly, \cite{gocmen_transportation_2019} used clustering method and upstream 3D-bin packing to provide context for VRP problem.

\section{ML-Assisted VRP Algorithms} \label{sec4}
Machine learning can help address combinatorial optimisation problems in various ways (see \cite{bengio_machine_2020} for a more general discussion or \cite{Smith1999} for a survey of older work).
It can assist traditional solvers that utilise heuristics to make decisions.
Those heuristics are created by experts, are generally hard to design, and may not transfer well from one problem (instance) to another.
Machine learning can help learn such heuristics automatically.
For instance, \cite{he_learning_2014} and \cite{khalil_learning_2016} independently proposed learning searching in a branch-and-bound tree for mixed-integer linear programs (MILP).
\cite{li_combinatorial_2018} trained a graph convolutional network to guide a tree search, notably for the Boolean satisfiability (SAT) problem.
Such  approaches are useful as they can benefit those aiming to solve problems that can be formulated as MILP or SAT. See \cite{lodi_learning_2017} for a recent survey on machine learning applied to branch and bound.
In this section, we review various strategies that combine machine learning with traditional VRP solution methods. 
    \subsection{ML-Guided VRP Decomposition Strategies} 
As a classic NP-hard combinatorial optimisation problem, VRP's search space grows exponentially with respect to the problem size, causing significant challenges to solve to optimality as the problem size increases. A common strategy is to decompose the original large scale problem into a number of smaller sub-problems. However, how the problem is decomposed becomes another difficult problem. In this section, we review decomposition based VRP studies that adopt machine learning to guide the decomposition.  

In general, in a decomposition approach, the VRP problem is divided into smaller and simpler sub-problems and a specific solution method is applied to each sub-problem \citep{archetti_survey_2014}. We classify the ML-guided VRP decomposition into \textbf{hierarchical} and \textbf{integrated} approaches. Hierarchical approaches are high-level combinations of ML and OR algorithms (e.g. either exact methods or (meta-)heuristics), while the integrated approaches are low-level combinations because one algorithm is embedded within another.

\subsubsection{Hierarchical approaches}
The ML-guided hierarchical decomposition in VRP can be classified into two types: Route First and Cluster Second (RFCS) and Cluster First and Route Second (CFRS). RFCS decomposes a VRP problem into two sub-problems: 1) Generate a TSP tour from the depot around all the customers and back to the depot; 2) Partition the TSP tour into a set of vehicle routes. Conversely, the CFRS starting from assigning customers to vehicles, then vehicle tours are constructed.

The OR approaches of solving RFCS (e.g. \cite{beasley_route_1983}, \cite{montoya_route-first_2014}) and CFRS (e.g. \cite{fisher_generalized_1981}, \cite{dondo_cluster-based_2007}) have been applied to many VRP problems. 

\textbf{ML-guided CFRS}
\cite{erdogan_green_2012} solved the Green Vehicle Routing Problem (G-VRP) using Modified Clarke and Wright Savings (MCWS) heuristic and an unsupervised clustering algorithm, which is built on the concepts from the Density-Based Spatial Clustering of Applications with Noise (DBSCAN) algorithm for exploiting the spatial properties of the G-VRP. The VRP is decomposed into two sub-problems: 1) Cluster customers using DBSCAN; 2) run MCWS to construct vehicle tours.

\cite{comert_new_2017} proposed a two-stage solution method for the Vehicle Routing Problem with Time Windows (VRPTW).  In the first stage, customers were assigned to vehicles using the best-performing clustering algorithms among K-means, K-medoids, and DBSCAN. In the second stage, a VRPTW was solved using Linear Programming (LP) to construct vehicle tours.

\cite{gocmen_transportation_2019} studied intermodal network problems combining pick-up routing problems with three-dimensional loading constraints, clustered backhauls at the operational level, and train loading at a tactical level. The authors first used the clustering of backhauls, then packed using K-means for a feasible loading pattern with four loading dimensions, and finally utilised capacitated VRP formulation which can be solved to optimally for  small problem instances. The K-means clustering ensures the assignments of tasks into several clusters, where the initial cluster number is defined before the solution procedure starts. The task assignment to which of the clusters is not part of the decision variables but is an input to the intermodal network problem.

The scientific literature shows that the ML-guided CFRS follows a similar solution procedure where an unsupervised cluster algorithm is first adopted to cluster customers to vehicles, followed by vehicle tours construction using (meta)-heuristics or LP methods. Readers are referred to \cite{he_balanced_2009}, \cite{nallusamy_optimization_2010}, \cite{korayem_using_2015}, \cite{reed_ant_2014}, \cite{comert_cluster_2018}, \cite{xu_dynamic_2018}, \cite{geetha_nested_2013}, \cite{rautela_distribution_2019}, \cite{geetha_metaheuristic_2012}, \cite{yucenur_new_2011}, \cite{qi_vehicle_2011}, \cite{luo_multi-phase_2014}, \cite{gao_ant_2016}, \cite{miranda-bront_cluster-first_2017} for similar and more complete references.

\textbf{ML guided RFCS}
We found only one study using ML in RFCS. \cite{kubra_route_2019} investigated the problem of determining service routes for the staff of a company. The authors first implemented a GA to create a TSP tour, and then partitioned and compared the tours by considering different scenarios using clustering methods including K-means, K-medoids and K-modes.

\subsubsection{Integrated approaches}
Integrated approaches of the decomposition have been applied to enhance the performance of LP algorithms. For example, the approximation of strong branching \citep{alvarez_machine_2017}, learning techniques introduced in the branch-and-bound algorithm \citep{lodi_learning_2017}. A recent survey of utilising ML to solve combinatorial optimisation problems can be found in \cite{bengio_machine_2020}. However, the integrated approaches tailored to the VRP are somewhat limited. 

\cite{desaulniers_machine-learning-based_2020} employed a Graph Neural Networks (GNN) to select a subset of the columns generated at each iteration of the column generation process. The input of the GNN is a set of promising columns generated by solving an LP model to minimise the number of columns and ensure a maximal decrease of the objective value. The learned GNN model is to reduce the computational time spent re-optimising the restricted master problem (RMP) at each iteration by selecting the most promising columns.  Computational results on two types of VRP indicate that average computational time reductions of 20 to 30\% are achievable when solving the RMP. 

\cite{salvagnin_learning_2017} applied a K-nearest-neighbour classifier to decide whether or not a Dantzig-Wolfe decomposition should be applied to a given problem, and which decomposition to choose. This method is not limited to VRP though. 

\cite{yao_admm-based_2019} proposed a decomposition framework based on Alternating Direction Method of Multipliers (ADMM) to iteratively improve the primal and dual solution quality simultaneously. ADMM has been utilised in distributed convex optimisation problems arising in statistics and machine learning (\cite{boyd_distributed_2010}). The authors first constructed a multi-dimensional commodity flow formulation for the VRP. Then, ADMM was applied to develop a decomposition framework, in which the original model was decomposed into a series of least-cost path problems which can be solved by the dynamic programming.

    \subsection{ML-Guided Perturbative VRP Algorithms} 
        When VRPs are modelled as offline optimisation problems (i.e. the problem related parameters are known and given prior to problem solving), perturbation based metaheuristics and evolutionary algorithms are among the most popular solution methods. The simplest perturbation search method is the basic local search. It searches a predefined neighbourhood of candidate solutions to find solutions which are superior to the current ones according to the objective function. The neighbourhood of a solution is defined as the set of solutions that can be derived by perturbing the incumbent solution according to certain transition rules. For example, 2-opt neighbourhood function for VRP operates in the neighbourhood of solutions solutions that only differ in the order of two connecting arcs. Since global search is generally impractical for NP-hard problems, local search can provide satisfactory, although not optimal, solutions within a reasonable time when well designed. See \cite{gendreau_handbook_2010} for more detailed discussions of meta-heuristics and local search algorithms. 
   
Local search starts from an initial solution (either randomly generated or heuristically constructed), and moves to a neighbour solution better than the current one continuously until no improvement can be made or the stopping criteria are met. Local search may be trapped in local optima. To overcome this shortcoming, a number of meta-heuristics approaches are introduced to improve the neighbourhood search strategies, for example simulated annealing, tabu search, guided neighbourhood search, variable neighbourhood search and adaptive large neighbourhood search. The idea of getting away from local optimum is to introduce a perturbation so that the search process can accept inferior solutions and ‘jump’ out poor local optima by following some guiding strategies. 

So far, ML has been used in different parts of perturbation based search approaches to improve its performance: initial solution generation, adaptive selection of neighbourhoods at different search stages, generation of neighbourhood functions and solution evaluations. We review ML guided initial solution generation in Section \ref{sec4:ml-construct-vrp}. In the following we focus on the intelligent neighbourhood selection, neighbourhood function generation and solution evaluations assisted by machine learning. 



\subsubsection{Learning to Select Perturbation Heuristics}
The concept of using machine learning techniques to guide the neighbourhood search is well documented. Initially, various adaptive mechanisms in the principle of basic reinforcement learning are used in the form of perturbative hyper-heuristics \citep{burke_classification_2019} and adaptive large neighbourhood search (ALNS) \citep{ropke_adaptive_2006}. The main idea of perturbative hyper-heuristic is to intelligently select a set of perturbative low-level heuristics to adapt to different problem solving scenarios. As such, the learning based components are crucial parts of this type of hyper-heuristic methods. The literature of applying learning assisted perturbative hyper-heuristics for various combinatorial optimisation problems is rich. Here, we shall focus mainly on the papers that are developed for vehicle routing. 

\cite{bai_memory_2007} addressed a VRPTW problem with a hyper-heuristic method that uses the concept of reinforcement learning to guide the selection of low-level heuristics and simulated annealing as the perturbation acceptance criteria. The study investigated  how the memory length of the adaptive learning mechanism affects the performance of the algorithm. \cite{sabar_dynamic_2015} proposed a new hyper-heuristic method for two VRP problems by using an improved reward scheme (dynamic multiarmed bandit-extreme value-based reward) in the learning mechanism, coupled with gene expression programming for acceptance criteria generation. \cite{soria-alcaraz_methodology_2017} extended this learning reward scheme with additional information from non-parametric statistics and fitness landscape measurements. 

\cite{garrido_dvrp_2010} investigated evolutionary-based hyper-heuristic approaches for a dynamic vehicle routing problem. The results showed that evolutionary based learning mechanisms improve the algorithm performance by adapting to different dynamic environments. 
\cite{asta_apprenticeship_2014} proposed to use an apprenticeship learning in the hyper-heuristics for vehicle routing. The trained algorithm is able to produce high quality solutions for test instances which are not seen during the hyper-heuristic training stage. 

The learning assisted perturbative hyper-heuristics have also been used in other VRP variants, including railway maintenance service routing \citep{pour_choice_2018}, multi-depot m-TSP problem \citep{pandiri_hyper-heuristic_2018}, multi-objective routing planning \citep{yao_parallel_2018}, mixed-shift full truckload routing \citep{chen_hyper-heuristic_2018,chen_variable_2020}, energy-aware routing \citep{leng_novel_2019}, urban transport routing \citep{ahmed_solving_2019,heyken_soares_public_2020}. 

One emerging direction for perturbative heuristic hyper-heuristics is the use of deep reinforcement learning (DRL) for heuristics selection. A DRL combines deep learning with aforementioned reinforcement learning. \cite{tyasnurita_learning_2017} used a time delay neural network (TDNN) as a classifier to select the low-level heuristics to solve open vehicle routing problem. Parameters of TNDD were trained through the experience replay from an ‘expert’ hyper-heuristic algorithm called MCF-AM (Modified Choice Function - All Moves) from \cite{drake_improved_2012}. 
\cite{chen_learning_2019} proposed a general deep reinforcement learning based hyper-heuristic framework for combinatorial optimisation problems called local rewriting framework (Neuwriter) which generates a solution iteratively. In this work, a region picker and a rule picker were defined and trained separately. At each rewrite step, two pickers selected a sub-area of current solution and its rewrite rule one behind the other. Then the modified sub-area will be put back to generate a new solution. The model repeated this process until convergence. The online job scheduling problem, expression simplification problem, and capacitated vehicle routing problem were used to test this method. A bi-directional LSTM layer was used to embed the input nodes in CVRP problem, and a similar pointer network mechanism was used to select a node through a probability distribution. \cite{wu_learning_2020} extended the Neuwriter framework by integrating the region picker and rule picker policy networks into one. Specifically, the compatibility computation was adopted in the model to produce a probability matrix of node pairs whose element specified the two corresponding nodes. In order to capture the node position information, sinusoidal positional encoding was introduced to the embedding layer. The results showed that this method outperformed Neuwriter. 

\cite{lu_learning-based_2020} proposed an iterative improvement method called `Learn to Improve (L2I)' based on that. Notably, this method outperformed LKH3 \citep{helsgaun2017extension} which was the state-of-the-art method of VRP on both speed and solution quality. The model started with a random initial solution and used two classes of predefined low-level heuristics, namely, improvement operators and perturbation operators which were used to local search the solution and destroy part of the solution respectively. A double layer controller guided by reinforcement learning was used to select the corresponding heuristics. When generating the action, the model would also consider the history actions and their effects. A policy ensemble mechanism was used to improve the generalisation.

\subsubsection{Learning to Adapt Neighbourhood Choices}
 Most real-life combinatorial optimisation problems involve complex objectives and constraints which often lead to very different solution space landscapes. For example, some solution spaces are very rugged with a lot of local optima while some other solution spaces contain plateaus that are insensitive to neighbourhood moves. These challenges lead to considerable research efforts in embedding learning mechanisms for more efficient neighbourhood search through adaptive neighbourhood selection. Since the early work on adaptive large neighbourhood search (ALNS) from \citep{ropke_adaptive_2006}, a number of follow-up research efforts have been made in either directly applying or extending the methods to solve different VRP variants. Papers that used similar algorithms for different VRP variants include \cite{laporte_adaptive_2010,ribeiro_adaptive_2012,kovacs_adaptive_2012,hemmelmayr_adaptive_2012,demir_adaptive_2012,masson_adaptive_2013,azi_adaptive_2014,aksen_adaptive_2014,belo-filho_adaptive_2015}. 
 
 There have been some research to further enhance ALNS method by hybridising with other methods. \cite{qu_grasp_2012} combined ALNS with a GRASP approach in solving a pickup and delivery problem with transshipment. \cite{parragh_branch-and-price_2017} combined a branch and price method with the ALNS method for a truck and trailer routing problem considering the time window constraints. \cite{zulj_hybrid_2018} integrated tabu search in a ALNS framework for solving an order-batching problem. \cite{lahyani_hybrid_2019} used a hybrid ALNS method to successfully address a multi-depot open vehicle routing problem. \cite{ha_new_2020} combined constraint programming with ALNS to solve VRP with synchronisation constraints. 
 
 Another learning-based ALNS algorithm proposed by \cite{hottung2019neural} was called neural large neighborhood search (NLNS). The method was specifically adapted to support parallel computing, which is one of the contributions of this method. Such features could support two implementation patterns: batch search which solved a set of instances simultaneously and the single instance search which solved only one instance concurrently. The potential of the method was demonstrated through experiments for capacitated vehicle routing problem (CVRP) and the split delivery vehicle routing problem (SDVRP).

\subsubsection{Learning to generate perturbation heuristics} 
A new exciting direction of using machine learning in solving VRP problems is that machine-learning models could be used to generate new heuristics to perturb solutions as apposed to using manually-crafted perturbation heuristics. For example, a neural network model takes an incumbent solution as inputs and outputs the indices of nodes that are modified. In this case, the neural network model itself acts as perturbative heuristic in the traditional optimisation algorithm. 

\cite{da_costa_learning_2020} focused on improvement (or perturbative) heuristics that could refine a given solution iteratively until reaching a near-optimal solution. In his work, a method that could learn a policy to generate the 2-opt heuristic for TSP was proposed. A similar encoder-decoder framework was used to encode the graph and generate a sequence of action distribution but the mechanism was modified to make it easy to extend to k-opt operations. The difference to the initial pointer network is that at each decoder step, the model outputs an action (two nodes for 2-opt) rather than one specific node to construct the final solution. 
 
\cite{gao_learn_2020} proposed a method for learning the local search heuristics. Similarly, an encoder-decoder was used and trained by actor-critic algorithm. Inspired by the ALNS algorithm for vehicle routing problems, the authors defined the local search heuristic as a destroy operator and a repair operator. The destroy operator removed a subset node of the current solution and the repair operator was used to generate a permutation of the selected elements and insert them back. Motivated by the graph attention network (GAT) mechanism \citep{velickovic_graph_2018} which is an effective method to represent the graph topology by propagating the neighbour node information through the attention mechanism, the authors proposed a modified version called Element-wise GAT with Edge-embedding (EGATE) which not only considered the information of nodes but also the arc between the nodes. The attention mask was generated by softmax the concatenation of embedding of arc and two nodes it connected with. The decoder acted as destroy and repair operation.

\cite{chen_dynamic_2020} were also motivated by ALNS algorithm, and proposed a similar method called dynamic particle removal (DPR) using Hierarchical Recurrent Graph Convolutional Network (HRGCN). Similar to the method of \cite{gao_learn_2020}, the authors also defined the local search as a destroy and repair operator. The degree (size and the allocation of the sub-nodes) of the destroy operator was dynamically determined. The HRGCN is able to be aware of spatial (graph topology) and temporal (embedding in previous iterations) context information. 

\subsubsection{Learning to Speedup Solution Evaluations} 
For many real world scheduling and routing problems, computing evaluation functions is expensive. ML has been used in evaluating solutions to reduce computational complexity. \cite{boyan_learning_2000} developed a general algorithm (which was called STAGE) to learn an evaluation function that predicted the outcome of a local search. The learned evaluation function was then used to guide the future search trajectories toward better optima on the same problem. \cite{moll_learning_1998} introduced an offline reinforcement learning phase to STAGE and compared using of learned evaluation function with the original evaluation function. The proposed algorithm was applied to the Dial-A-Ride Problem, a variant of TSP, to show how well learning an instance-independent evaluation function could guide local search for additional instances.

    \subsection{Learning to Construct VRP Solutions} \label{sec4:ml-construct-vrp} 
            In this section we discuss some representative works that exploit machine learning in the constructive approach for vehicle routing related problems.
    Recall that such an approach consists of building iteratively a complete solution from scratch (e.g., in TSP, it sequentially selects unvisited cities until a tour is formed) in contrast to approaches based on iterative perturbative search.

    
    The most common approach is to learn a probability distribution over solutions, which can then guide a tree search (e.g., greedy search, beam search) to generate a full solution.
    In contrast, another possibility is to learn directly a constructive solver.
    Indeed, such a solver can be seen as a sequential decision-maker, which can be trained via an evolutionary or a reinforcement learning process.
    
    Using machine learning to solve routing problems has a long history (e.g., \cite{hopfield_neural_1985,potvin1990integration}).
    We refer the reader to other surveys on historical developments (e.g., \cite{Smith1999}).
    We will focus mostly on more recent applications of machine learning to these routing problems.
    We discuss first the works that focus on (mostly planar) TSP or variants, and then turn to those on VRP and its variants.
    We mention work on online VRP and variants at the end.




    For TSP, \citet{vinyals_pointer_2015} proposed Pointer Networks, which is a novel neural network architecture with two components.
    First, an encoder was implemented as a recurrent neural network sequentially reading the positions of the cities.
    Second, a decoder was realized as a recurrent neural network iteratively outputting a probability distribution over remaining cities, using an attention mechanism \citep{Bahdanau2015}.
    The whole model was trained in a supervised fashion.
    For testing, beam search was used to ensure that only valid tours are output.
    The technique was applied to TSP instances with up to 50 nodes.
    
    \citet{bello_neural_2017} extended the previous approach to train the network using reinforcement learning with an actor-critic scheme.
    They used the cost length of a generated tour as an unbiased estimate of the value of a policy.
    The authors found out that pretraining on a set of instances in addition to training on the particular instance to be solved yielded the best performance on TSP instances with up to 100 nodes.
    Some more recent investigation \citep{joshi_learning_2019} suggests that RL may lead to better generalization capability compared to supervised learning.
    
    \citet{khalil_learning_2017} proposed a general method to solve graph-based combinatorial optimization problems based on graph embedding to encode a partial solution and reinforcement learning to learn a greedy policy. 
    In contrast to the previous approaches, they used a value-based RL method, fitted Q-learning. 
    
    \cite{deudon_learning_2018} proposed a graph attention network architecture to improve over Pointer Networks to obtain invariance over input order.
    Also, the authors introduced the idea of preprocessing the inputs with PCA to obtain rotation invariance.
    With a critic using a similar architecture, the policy was trained in an actor-critic architecture.
    They used a mask to remove already visited cities.
    In their recent hybrid method, the authors proposed further to improve the solutions sampled from the trained policy using the 2-OPT heuristics.
    
    In contrast to other approaches mentioned here, 
    \citet{nowak_note_2017} proposed a non-autoregressive model based on graph neural networks, i.e., their model does not select cities sequentially.
    Instead, the model trained in a supervised way outputs an adjacency matrix representing a distribution over tours, from which they extract a full solution via beam search.
    Although the method is not competitive with other deep learning methods, 
    \citet{joshi_efficient_2019} improved this approach by using notably graph convolutional networks.
    They showed that their approach compares favourably with other previous autoregressive approaches.
    
    \citet{yang_boosting_2018} proposed to perform the dynamic programming update of the Bellman-Held-Karp algorithm \citep{Bellman1962,Held1962} for solving TSP in an approximate way using neural networks.
    Doing so allows their proposition to tackle much larger TSP instances.
    
    \cite{ma_combinatorial_2020} proposed Graph Pointer Network (GPN), an extension of Pointer Network, with a graph embedding of the input.
    They demonstrated promising generalisability of GPN, which can be trained on small TSP instances (up to 100 cities) and then solve larger instances up to 1000 cities.
    Besides, they suggest using a hierarchy of GPNs to take into account additional constraints on the TSP problem.
    They tried their ideas on TSP with time windows and showed that their approach performs well.
    
    Some works considered the multiple TSP (mTSP), where several traveling salesmen need to visit all the cities exactly once in a cooperative manner.
    This problem can be seen as a relaxed VRP problem.
    \citet{kaempfer_learning_2019} adapted PointNet \citep{Qi2017}, which deals with sets of points, to this mTSP problem. 
    \citet{hu_reinforcement_2020} solved it by first using  a cooperative multi-agent deep reinforcement learning for agent-to-city assignment, and then computing the tour of each agent using a classic TSP solver.
    
    While the previous works only focus on TSP, recent works started to consider the harder problems of VRPs and variants.
    \citet{nazari_reinforcement_2018} proposed a simplified version of Pointer Networks to solve capacitated VRPs, split-delivery VRPs (SDVRP) and stochastic VRPs. 
    In order to make the input invariant to sequence order (e.g., order of customers), they replaced the RNN encoder of Pointer Networks by simple embedding maps. 
    The resulting model can then handle changes in the input (e.g., customer demand after being visited).
    An actor-critic scheme was used for training and beam search, which tracks the most probable paths, was used to generate the final best solution.
    
    \citet{kool_attention_2019} proposed a transformer-based model to solve routing problems, such as TSP, CVRP, or split delivery VRP.
    The model is similar to that of  \citet{deudon_learning_2018} with a few simplifications and improvements.
    In particular, in contrast to that work, \citet{kool_attention_2019} did not use 2-OPT.
    Another novelty is that policy gradient with a self-critic baseline (estimated with greedy policy rollouts) was used for training. 
    \cite{peng_deep_2020} generalized this approach to use a dynamic attention model so that state features can be updated during the construction of a solution.
    
    \citet{sheng_pointer_2020} proposed a variation of Pointer Network to solve VRP with Task Priority and Limited Resources.
    The model was trained in the RL setting.
    They showed that this approach is comparable to Genetic Algorithm (GA) for medium-sized instances ($\approx 50$ cities), but its performance becomes better than GA for larger-sized instances ($>100$ cities) while taking much less computational time. 
    
    \citet{duan_efficiently_2020} proposed a technique that combines training with reinforcement learning and supervised learning.
    The method was based on graph convolutional network to encode a problem instance with node and edge features.
    Node features were used as input of an RNN policy to output a solution, which was used to train a classifier taking edge features as inputs to predict the probability of selecting an edge in a solution.
    The method was evaluated on real-world data sets and shown to generalize well.

    Apart from the work by \citet{nazari_reinforcement_2018}, who considered SVRP, there is scarce  literature on dynamic and stochastic VRP using modern machine learning techniques.
    One exception is \citet{balaji_orl_2019} that considers stochastic and dynamic capacitated VRP problems with pickup and delivery, time windows and service guarantee.
    The authors showed that deep RL algorithms can directly be trained to solve them and they are competitive or superior to classic baselines.

    On the other hand, traditional evolutionary algorithms have advantages in dynamic capacitated VRP problems, especially with the assistance of heuristic-based methods. \cite{sabar_grammatical_2013} investigated a hyper-heuristic method assisted by a grammatical evolutionary method for capacitated VRP problem.
        
    \cite{jacobsen-grocott_evolving_2017} attempted to use a hyper-heuristic method to solve Dynamic Vehicle Routing with Time Windows (DVRPTW). Such problems require the acceptance or rejection decisions for dynamically arriving customer requests. The genetic programming evolved heuristics were used to determine whether or not to accept new requests and add them to the current routes. The results showed that with the dynamism degree increasing, the GP-evolved heuristics significantly outperformed the handcrafted heuristics.
    
    \cite{liu2017automated} developed a new Genetic Programming-based Hyper-Heuristic (GPHH) for automated heuristic design for Uncertain Capacitated Arc Routing Problem (UCARP), and designed a novel effective meta-algorithm. Their experimental results showed that the proposed GPHH significantly outperforms the existing GPHH methods and manually designed heuristics.
    
    To gain more interpretable routing policies, \cite{wang_evolving_2019} used three ensemble genetic programming methods, namely, Bagging GP, Boosting GP, and Cooperative Co-evolution GP, to solve uncertain capacitated arc routing problem. Evolved depth-limited tree expressions correspond to an ensemble of these methods to represent the priorities of each task in an instance. The results showed that an ensemble of simple policies is able to compete with the complex policies while maintaining their high interpretability. 
    
    \cite{maclachlan_genetic_2020} proposed to evolve collaborative routing policies within a data-driven genetic programming hyper-heuristic algorithm for a capacitated arc routing problem with uncertainties.  
    
    On the application side, \cite{chen2020data} introduced genetic programming based hyper-heuristic method to solve  a realistic VRP in marine container port with various uncertainty. The results confirmed the effectiveness of GP for this kind of dynamic uncertain problem.
    
\section{Concluding Remarks and Future Directions} \label{sec5}      
The vehicle routing problem and its variants are one of the most studied combinatorial optimisation problems in the research community because of its close relevance to transportation problems in industrial and societal activities, and yet few known algorithms have solved them to a satisfactory level. The main challenges lie in the scale of  real-world problems, complexity in objectives and constraints (non-linearity, dynamic nature) and uncertainties.  This literature review focuses on research efforts in integrating data analytics and machine learning in addressing these challenging VRP problems. A number of observations can be made.

\subsection{Problem diversity and dynamics}  
One of most challenging aspects of VRP is its diverse and dynamic nature. As indicated in \cite{eksioglu_vehicle_2009}, the varieties of VRP can be attributed to the diverse problem scenario characteristics, its physical characteristics, as well as its information characteristics. This leads to numerous VRP models that have been developed to capture the main structures of various real-world problems. Although some research efforts have been made to generalise these models, big challenges still exist in terms of solution methods because these generalised models may not be able to take advantage of the underlying special structures that can be exploited for the development of more efficient algorithms. Therefore, a growing research direction is to automate (at least partially) the modelling process of real-life VRP problems by integrating  data analytics and machine learning methods so that the key patterns and structures of the problems can be automatically identified and the most suitable VRP variants can be matched to the problem at hand. For practical applications, it, therefore, makes sense to develop a VRP expert system with a repository of parameterized VRP models and their corresponding algorithms so that the real-world problems can be automatically analysed, clustered and matched with one of existing model-algorithm pairs and readily solved. 

Nevertheless, one should also recognise the dynamic nature of the VRP problems in real life. The problem can diverge from one variant or type to another over time. Therefore, such an expert system must be adaptive to the dynamic changes of the environments, leading to the requirements of using more advanced AI methods so that the system can evolve and improve automatically. Therefore, the ability to self-learn and self-evolve over time will be a key feature of the next generation VRP expert systems. 

\subsection{Perturbative improvements vs. generative approaches}
As a classic NP-hard problem, VRP problems are often solved heuristically via iterative procedures, which can be broadly categorised in two different ways, namely perturbative and generative. A perturbative method is a kind of create-improve process that assumes a deterministic (i.e. offline) model and aims to improve the quality of the solution incrementally while having ability to escaping from poor local optima. Various learning mechanisms can be used to exploit the structures of the solution landscapes. Machine learning has proved to be beneficial in helping select among a set of pre-defined perturbative heuristics or neighbourhoods in the most appropriate way. It can also be used for the automation at lower level. For example, ML can generate perturbation heuristics automatically. It can also be used to approximate the expensive solution evaluations to speed up the local search process. 

Generative VRP approaches seek to build a high quality solution from scratch. Because the training process is often done offline, this type of methods has advantages compared with perturbative algorithms in terms of solution time and ability to handle uncertainties and problem related dynamics. Its main weakness is the quality of the resulting solutions which are often inferior compared with those from perturbative methods. However, with the advances in the computing power and deep learning (in particular the deep reinforcement learning), generative VRP algorithms have gained increasing popularity in recent years and have achieved more successes in solving practical VRP problems.  

\subsection{Model driven vs. data driven}
Although in the early stage of the VRP research, significant research attentions have been paid on the perturbative methods that strive to search for the global optimality to a given problem formulation. However, it is increasingly recognised that, due to the problem uncertainties and dynamics, the perceived optimality is very rarely realised in practice. It is often impractical (if ever possible) to formulate the real-life VRP problems exactly because the resulting models will most likely be intractable. A compromise has to be made between the accuracy of the models and their tractability. This can be very challenging for practitioners. 

As an alternative, data driven methods have been proposed as end-to-end solvers to VRP problems to reduce (or remove) the requirement of mathematical models. These methods train deep neural networks to produce solutions directly without the mathematical formulations, potentially making them easier to be applied in the real world. Therefore, these data driven methods often serve as black-box solvers and are often criticised for their poor interpretation abilities. Another stream of data-driven methods are based on the principles of genetic programming that evolves decision trees for solution constructions for VRP problems based on historical data. When suitable constraints and requirements are enforced during the evolution process, the resulting GP trees tend to have better interpretation abilities than the solutions from the deep neural networks.  

\subsection{Challenges and Prospects} \label{sec6}
Despite the fast growing popularity of integrating ML in VRP research, the research community still faces a number of challenges. 

Firstly, the VRP is often part of a larger complex system that requires a good level of reliability or robustness across all possible scenarios to ensure the system does not reach certain undesirable (or disastrous) states. In addition, a good level of interpretability of the adopted algorithms is also required. Although there has been some research efforts in the area of explainable AI and verification methods, more theoretic research must be conducted to lay solid foundations for the future research and broader applications. 

Secondly, there lacks a high-quality ML-assisted VRP research platform for training and testing purposes. The required resources and libraries are very scattered in two different research communities in machine learning and operations research, respectively. To help grow the ML-assisted VRP research, the two research communities must work together towards an integrated platform with libraries and tools for both ML and optimisation. Of course, this would also lead to opportunities for multidisciplinary research. 

Thirdly, many of the current ML-assisted VRP research requires huge amount of data which is normally not available directly. The trained models do not generalise well across different instances, scenarios and problem domains. This lack of data  and model generalisation is largely caused by the trial-and-error nature of the traditional deep reinforcement learning. Important information regarding the objective function(s) and constraints is not fully exploited. More novel approaches are required to provide a better fusion of the current analytical methods based on mathematical models and neural network methods driven by data. 

Last but not least, more real-world applications must be encouraged to address the criticisms of the existing VRP research. One must balance the adoption costs and the performance of the solution methods. In particular, a much more powerful VRP simulation software is required to help practitioners to build customised VRP  environment with high performance in speed at a low cost. In addition, the research community should also consider  building pre-trained VRP models/libraries to reduce the training costs further. 

\section*{Acknowledgement}
This work is supported by the National Natural Science Foundation of China (grant number 72071116, 71471092), the Zhejiang Natural Science Foundation (grant number LR17G010001) and the Ningbo Science and Technology Bureau (grant numbers 2019B10026, 2017D10034). 

\bibliographystyle{tfcad}
\bibliography{ref-for-all}
\end{document}